\documentclass[letterpaper, 10pt, conference]{ieeeconf}  

\IEEEoverridecommandlockouts                              

\usepackage{amsmath}
\DeclareMathOperator*{\argmin}{\arg\!\min}
\usepackage{algorithm}
\usepackage[noend]{algpseudocode}
\usepackage{graphicx}
\usepackage{stfloats}
\usepackage{subfigure}
\usepackage{caption}
\usepackage{setspace}
\usepackage[colorlinks,linkcolor=blue]{hyperref}

\title{\LARGE \bf
Learn by Observation: Imitation Learning for Drone Patrolling from Videos of A Human Navigator
}

\author{Yue Fan$^{1,2}$, Shilei Chu$^{1}$, Wei Zhang$^{1}$*, Ran Song$^{1}$*, and Yibin Li$^{1}$
\thanks{This work was funded by the National Natural Science Foundation of China under Grant 61991411 and Grant U1913204, the National Key Research and Development Plan of China under Grant 2017YFB1300205, the Shandong Major Scientific and Technological Innovation Project 2018CXGC1503, the Young Taishan Scholars Program of Shandong Province No. tsqn20190929, and the Qilu Young Scholars Program of Shandong University No. 31400082063101.}
\thanks{$^{1}$ School of Control Science and Engineering, Shandong University, Jinan, China.}
\thanks{$^{2}$ Laboratory for Computational Sensing and Robotics, Johns Hopkins University, Baltimore, Maryland 21218, USA}
\thanks{* Corresponding authors: Wei Zhang and Ran Song (\{davidzhang, ransong\}@sdu.edu.cn\})
}
}
\begin{document}



\maketitle
\thispagestyle{empty}
\pagestyle{empty}


\begin{abstract}
We present an imitation learning method for autonomous drone patrolling based only on raw videos. Different from previous methods, we propose to let the drone learn patrolling in the air by observing and imitating how a human navigator does it on the ground. The observation process enables the automatic collection and annotation of data using inter-frame geometric consistency, resulting in less manual effort and high accuracy. Then a newly designed neural network is trained based on the annotated data to predict appropriate directions and translations for the drone to patrol in a lane-keeping manner as humans. Our method allows the drone to fly at a high altitude with a broad view and low risk. It can also detect all accessible directions at crossroads and further carry out the integration of available user instructions and autonomous patrolling control commands. Extensive experiments are conducted to demonstrate the accuracy of the proposed imitating learning process as well as the reliability of the holistic system for autonomous drone navigation. The codes, datasets as well as video demonstrations are available at \url{https://vsislab.github.io/uavpatrol}.



\end{abstract}


\section{Introduction}
Observation is the most natural way for human beings to learn new knowledge. Through observation, a human can extract the policy behind a certain task, which turns to be the main idea of imitation learning (IL) for intelligent robots. However, an emerging problem is that some IL techniques require the demonstration information to include the demonstrator's action \cite{torabi2019recent}. For example, many existing techniques \cite{amini2019variational}\cite{codevilla2018end}\cite{pan2018agile} train autonomous driving networks with the data of how humans control cars instead of letting the robot learn from videos recording how cars behave in similar scenarios. The data they collected include not only videos but also control signals, which potentially limits the performance of the robot 
and also increases the manual effort.  

\begin{figure*}[t]
\hsize=\textwidth
    \setlength{\abovecaptionskip}{-0.1cm}
    \setlength{\belowcaptionskip}{-0.2cm}   
    \hspace{0.06cm}
    \subfigure[Inaccurate map] 
    {
        \centering
        \includegraphics[width=0.15\textwidth,height=4.2cm]{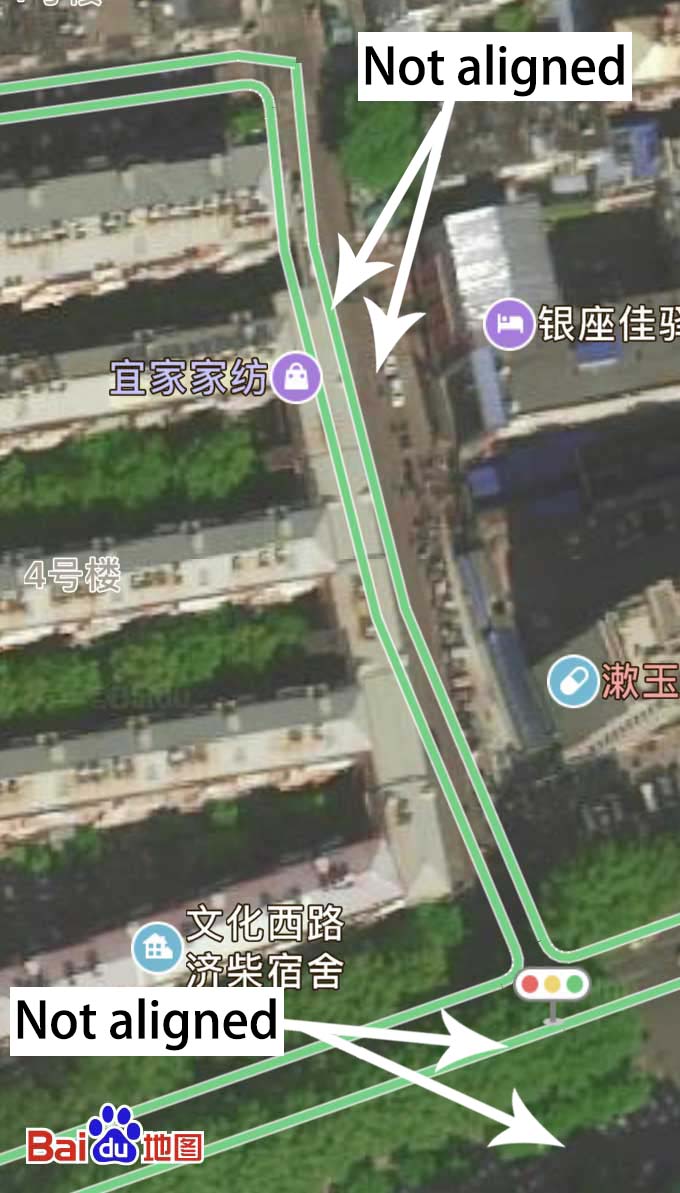}
        \includegraphics[width=0.15\textwidth,height=4.2cm]{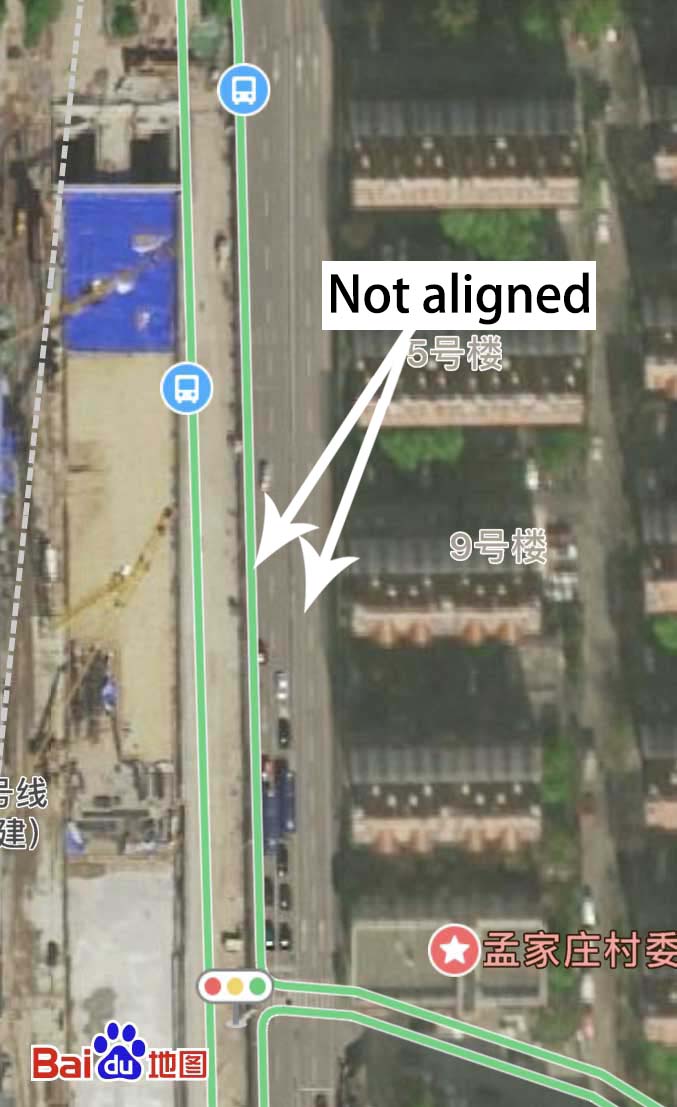}

        \label{fig:inaccurate}
    }
    \subfigure[Observations from drone's perspective]
    {
        \centering
        \begin{minipage}[b]{0.205\textwidth}
            \centering
            \includegraphics[width=1\linewidth,height=2.05cm]{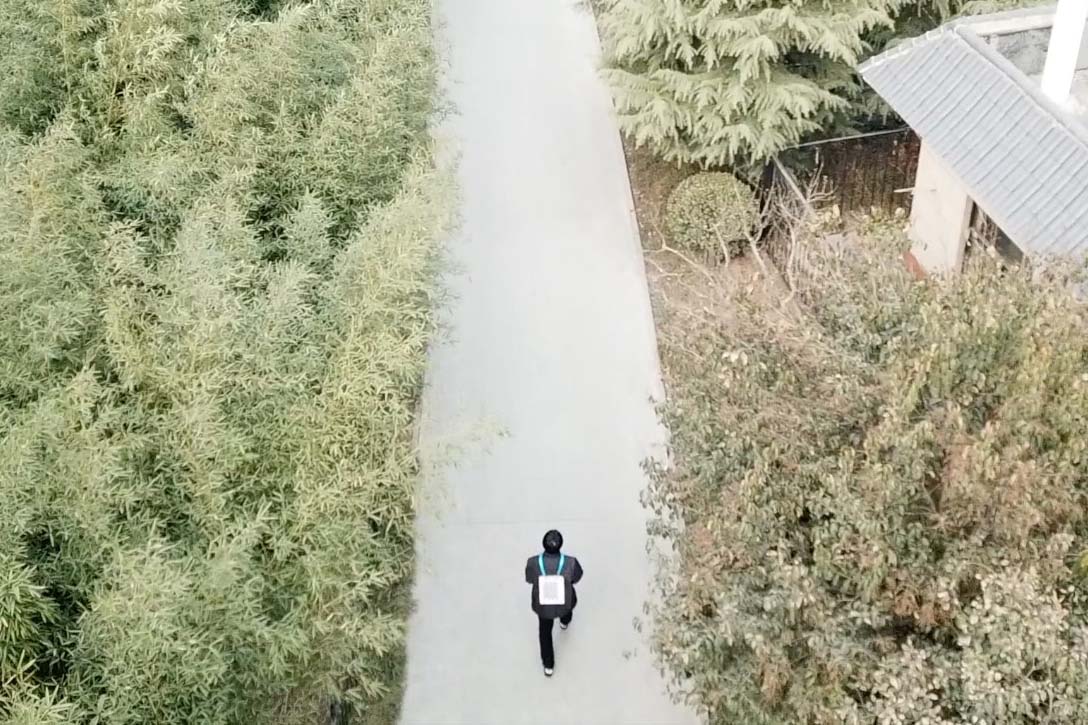}\vspace{0.1cm}
            \includegraphics[width=1\linewidth,height=2.05cm]{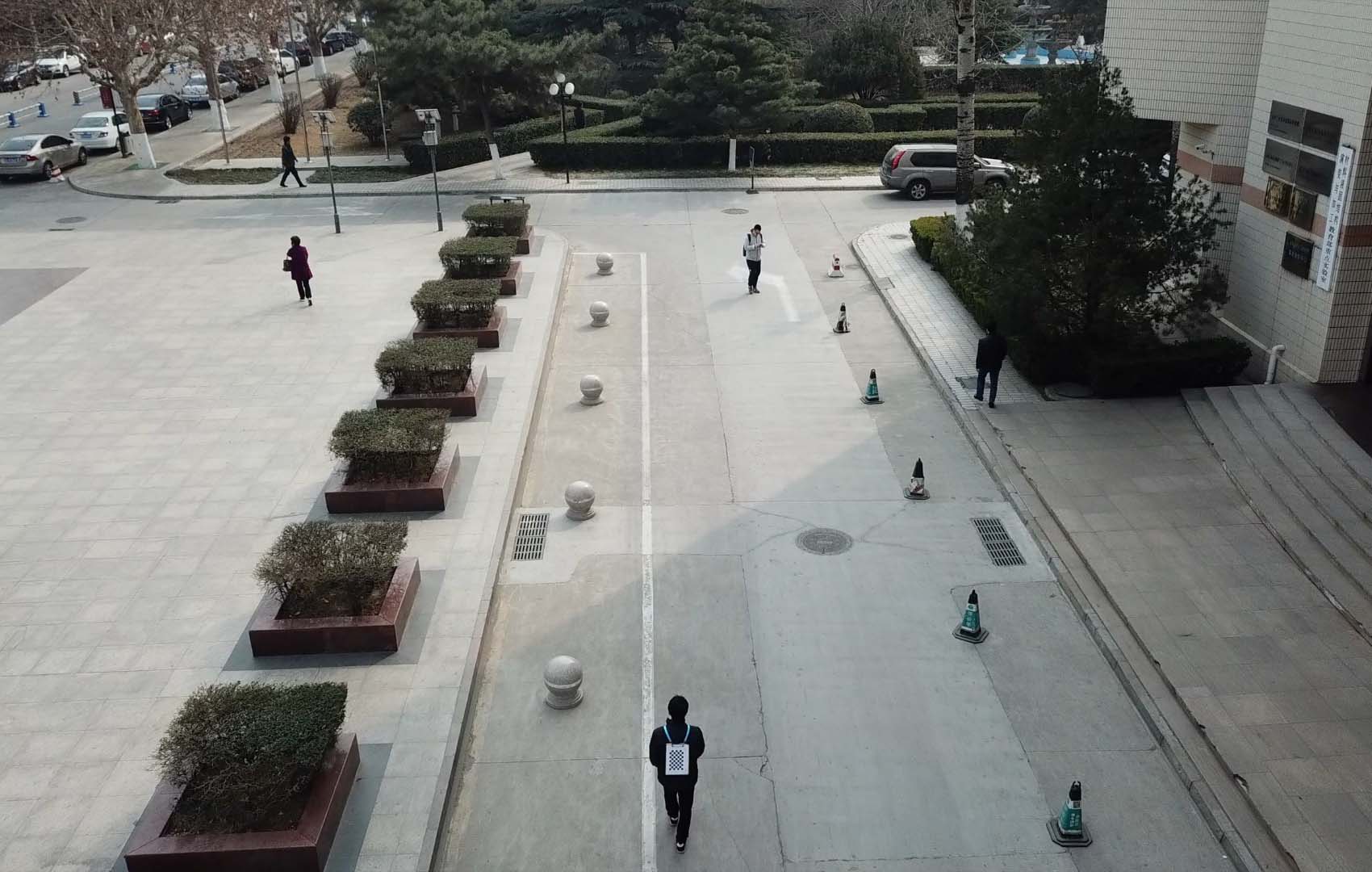}
        \end{minipage}
        \begin{minipage}[b]{0.205\textwidth}
            \centering
            \includegraphics[width=1\linewidth,height=2.05cm]{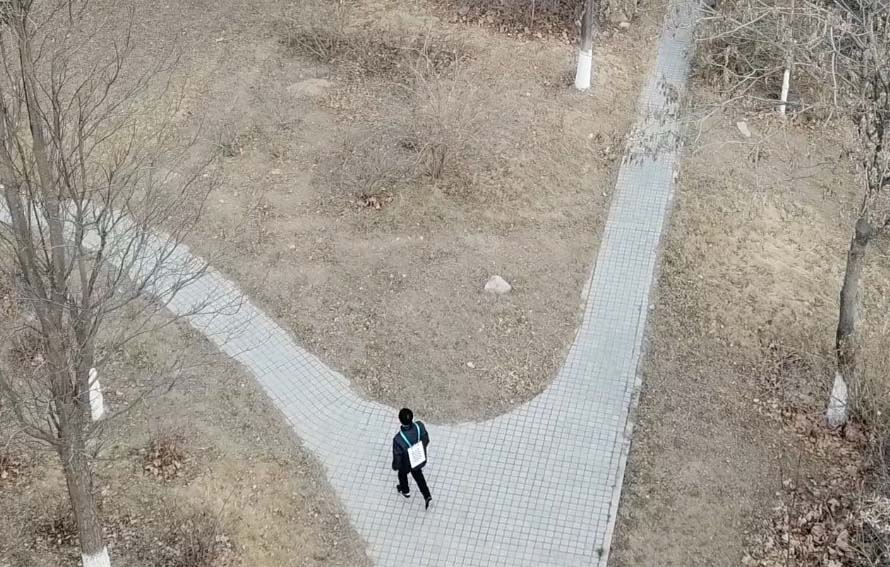}\vspace{0.1cm}
            \includegraphics[width=1\linewidth,height=2.05cm]{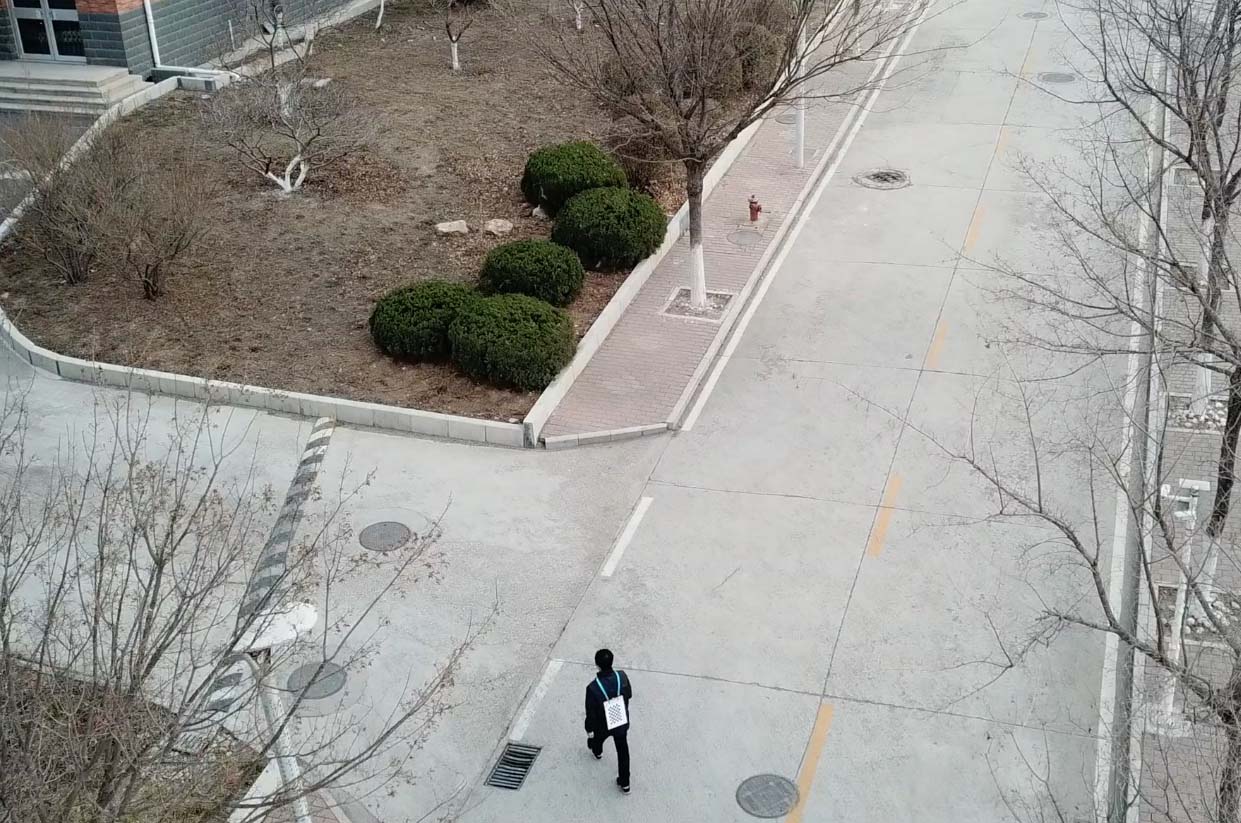}
        \end{minipage}
        \label{fig:imitation}
    }  
    \hspace{0.02cm}
    \subfigure[Observed navigator] 
    {
    \includegraphics[width=0.18\textwidth,height=4.2cm]{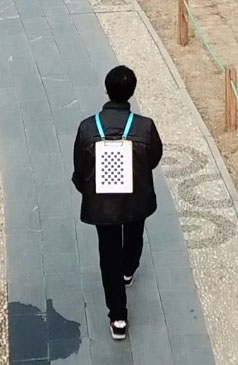}
    \label{fig:navigator}
    }
		\vspace{-1pt}
		\caption{
        \footnotesize{(a) In many cases, maps suffer from the undesired fact that the roads in a map are not perfectly aligned with the GPS data. (b) While tracking the navigator, drone observes the motion and learns the strategy of patrolling. The observation process involves various crossroad scenes. (c) What the drone observes in our dataset is mostly a person walking while having a calibration board on his back to help drone tracking.
        } 
    }
		\vspace{-2pt}
\end{figure*}

Inspired by \cite{liu2018imitation}\cite{sermanet2018time}\cite{yang2019learning} where robotic arms learned human behavior through IL, we intend to develop a method where the drone learns how to patrol like human from raw videos collected by its onboard camera. This is different from most of the existing methods for drone patrolling. For instance, some current works on drone patrolling rely heavily on the map \cite{lu2018survey}, which raises an issue: many maps are inaccurate for patrolling applications as the roads in the map are not perfectly aligned with the real roads as shown in Fig.~\ref{fig:inaccurate}
{due to measurement errors or national security restriction of maps in some countries}. Consequently, the drone is likely to fly off the right track. Some other drone navigation methods based on deep learning \cite{kim2015deep}\cite{loquercio2018dronet}\cite{smolyanskiy2017toward} do not need maps and train navigation neural networks with data collected from cameras mounted on cars or carried by people. As a result, these works are only suitable for drones that fly several meters off the ground and use forward-facing onboard cameras with limited views inadequate for autonomous patrolling. Also, they lack the ability to turn to the desired direction at intersections like in Fig.~\ref{fig:imitation}, and can only make the drone go along the direction with the minimum rotation angle.

In this work, we first present a method for automatically creating and annotating a dataset, namely Patrol Dataset that records human's patrolling behavior for IL. Usually, it is hard to collect and annotate a patrolling dataset by manually piloting the drone as the drone flying at a high altitude is sensitive to small changes of control commands and vulnerable to human mistakes. However, our automatic method by observing human patrol does not only create a dataset, but also converts the knowledge about human's patrolling behavior conveyed by the dataset to two annotations, i.e. the direction and translation of the human patrol. The drone tracks and takes videos of a human navigator patrolling along a road, as shown in Fig.~\ref{fig:imitation} and Fig.~\ref{fig:navigator} when flying at an altitude of tens of meters with the onboard camera downward. To annotate the video frames, we designed a frame-by-frame matching algorithm extracting geometry information between consecutive frames to automatically generate the annotations for each frame. In this way, the outcomes of drone observation, i.e. the raw videos, are converted to an annotated dataset to train a specifically designed neural network for learning how to patrol like humans.

The proposed IL network is named UAVPatrolNet which includes three sub-networks: Perception-Net, Direction-Net and Translation-Net. The drone first utilizes the Perception-Net to extract features from a single image. Then, instead of only constantly going along the direction with the minimum rotation angle at a crossroad, the drone relies on the Direction-Net to predict the probabilities of all possible road directions. Meanwhile, based on the deep features output by the Perception-Net, the Translation-Net learns the translations needed by the drone as a regression problem, which ensures that the drone can fly in a lane-keeping manner as humans. Finally, a controller takes as input the output of the UAVPatrolNet to send patrolling commands to the drone. It is worth mentioning that to accomplish user-defined tasks, the controller is designed to be compatible with user instructions derived from a user interface. It can fuse the commands yielded by the UAVPatrolNet with user instructions provided either before or during the flight via map annotations, buttons, or even speeches. 

To build such an autonomous drone navigation system based on IL, we develop both software and hardware tailored for the patrol task. The contribution of our work is threefold:
\begin{itemize}
\item A new scheme for the autonomous collection and annotation of the drone patrolling dataset, i.e. Patrol Dataset that enables the observation-based IL;
\item A new neural network composed of three sub-networks for IL that trains the drone to patrol like humans through the annotated dataset;
\item A newly designed user interface and flexible drone patrolling controller with the ability to fuse user instructions whenever available.
\end{itemize}

\section{Related Work}

The drone has gained enormous popularity recently and autonomous drone navigation is an active topic in the field. Many kinds of sensors are mounted to make a drone fly autonomously and avoid obstacles, such as ultrasound sensors \cite{guanglei2016application}\cite{paredes20183d}\cite{tarazona2014anti}, laser-scanners \cite{qin2016stereo}\cite{kownacki2016concept}\cite{klingbeil2014towards}, stereo cameras \cite{ramon2017detection}\cite{chen2017design}
or the combination of multiple sensors \cite{gageik2015obstacle}\cite{li2017small}. 
However, using sophisticated sensors imposes additional costs, and increases the load of the drone, which leads to shorter flying time. Thus recently, some researchers \cite{choi2016open}\cite{alvarez2016collision}\cite{mcguire2017efficient} only used a single onboard camera for drone navigation. Also, due to the lack of sufficient data and the risk of crashes, there is a trend to use simulation to train the navigation system. Chen et al. \cite{chen2018learning} collected a dense forest trail dataset in a simulated environment powered by Unreal Engine and used the simulated dataset to train a deep neural adaptation network. Fereshteh et al. \cite{chen2018learning} introduced deep reinforcement learning to train vision-based navigation policies and then transferred them into the real world. Kang et al. 
{Loquercio et al. \cite{loquercio2019deep} trained a racing drone using only simulated data.} However, one common problem of the simulation-based algorithms is that they need to take the risk of unsatisfied training results caused by the gap between the real-world and the simulated environments. 

Observation-based IL received much attention recently in robotics, especially in controlling robotic arm and other agents in simulated environments. It is a more direct and natural way of\hspace{-1pt} learning. \hspace{-1pt}Liu \hspace{-1pt}et \hspace{-1pt}al. \hspace{-1pt}\cite{liu2018imitation}\hspace{-1pt} realized\hspace{-1pt} imitation-from-observation for robot arm tasks such as block stacking. Sermanet et al. \cite{sermanet2018time} produced end-to-end self-supervised human behavior imitation without using any labels but only multi-view observation. Torabi et al. \cite{torabi2018behavioral} explored how to make autonomous agents in simulator learn the appropriate actions only from observation. However, as for autonomous moving robots such as cars and drones, most IL methods are not based purely on observation. The main reason is the difficulty of extracting information and forming a dataset only from observation. Amini et al. \cite{amini2019variational} and Codevilla et al. \cite{codevilla2018end} trained deep networks for autonomous car navigation on demonstrations of human driving, where the demonstration information includes the steering and throttle data. Loquercio et al. \cite{loquercio2018dronet} proposed a residual convolutional architecture generating safe drone flights in an urban environment, but the training dataset was a record of steering and braking commands when people drove cars and rode bicycles. Smolyanskiy et al. \cite{smolyanskiy2017toward} and Giusti et al. \cite{giusti2015machine} trained the deep network with only video information but their drone navigation datasets were created from videos taken from human perspective when walking, which is to some extent different from the actual drone flight where the drone should have its onboard camera facing downward when flying high. In this case, their methods could lead to difficult generalizing processes and stop the drone from flying at a high altitude as well.
By contrast, we develop an IL method to train the drone navigation system through the observation from the drone itself flying much higher with a downward onboard camera. 


\section{Proposed Method}
This section elaborates each of the three stages of the proposed method. In the first stage, we collect the video dataset which records human's patrolling behavior. To enable IL, we then propose an auto-labeling algorithm that converts the knowledge about human's patrolling behavior embedded in the dataset into annotations. In the second stage, the drone learns how to patrol like humans via the UAVPatrolNet using the annotated dataset. Finally, a controller is developed to output appropriate control commands based on the fusion of the output of the UAVPatrolNet and the optional user instructions for the automatic drone patrol.

\setcounter{figure}{1}

\renewcommand{\algorithmicrequire}{\textbf{Input:}}
\renewcommand{\algorithmicensure}{\textbf{Output:}}
\begin{algorithm}[t]
\caption{Auto-labeling algorithm}\label{euclid}
\begin{algorithmic}[1]
\Require $\newline Video Frames (\textit{I}_0 ... \textit{I}_{N-1}) \quad  // \text{Successive frames} \newline \quad  Initial Bounding Box(\textit{B}_0)\quad //\text{Defined by user}$
\Ensure $\newline Direction Labels (\textit{D}_0 ... \textit{D}_{N-2}), \newline Translation Labels (\textit{T}_0 ... \textit{T}_{N-2})$
\Function {GetHomography}{$\textit{I}_i, \textit{I}_{i+1}$}
\State $\textit{P}_i \gets \text{SIFT-Detector}( \textit{I}_i )$
\State $\textit{P}_{i+1} \gets \text{SIFT-Detector}( \textit{I}_{i+1} )$
\State $M \gets \textit{findGoodMatches}( \textit{P}_{i}, \textit{P}_{i+1} ) $
\State $M' \gets \text{RANSAC-Filter}( M )$
\State $\textit{H} \gets \text{findHomography}(M')$
\State \Return{$H$}
\EndFunction

\Function{Main}{$\textit{I}_0 ... \textit{I}_{N-1}, \textit{B}_0$}
\State $i \gets 0$
\While{$i+1<N$}
\State $\textit{B}_{i+1} \gets \textit{objectTracker}(\textit{B}_i, \textit{I}_i, \textit{I}_{i+1}) $
\State $\tilde{B}_{i+1} \gets \textit{getCenter}(\textit{B}_{i+1})$
\State $\tilde{B'}_{i+1} \gets \Call{GetHomography}{\textit{I}_{i+1},  \textit{I}_i}*\tilde{B}_{i+1}$
\State $\tilde{B}_{i} \gets \textit{getCenter}(\textit{B}_{i})$

\If {$|\tilde{B}_{i} - \tilde{B'}_{i+1}| > ReasonableDistance$} 
\State \Return {$Error$}
\EndIf
\State $\textit{D}_{i} \gets |\tilde{B}_{i} - \tilde{B'}_{i+1}|_{horizontal}/|\tilde{B}_{i} - \tilde{B'}_{i+1}|_{vertical}$

\State $\textit{T}_{i} \gets \tilde{B}_{i}.x+\textit{image}_{height}*\textit{D}_{i}(i)$.
\State $i \gets i+1$
\EndWhile
\EndFunction
\end{algorithmic}
\end{algorithm}

\subsection{Automatic Dataset Collection and Annotation}

The proposed Patrol Dataset is collected and annotated fully automatically by observing how the human navigator patrols along the road, as described in detail below:

{\it Dataset Collection:} Through the observation process, the data is first collected from the same perspective as drone patrolling, which to the best of our knowledge, is unique from all other available datasets. We use a DJI drone to record videos while it is tracking a person (the so-called {\it navigator}) that is moving forward along the road {with build-in tracking algorithm}. The tracking algorithm is a built-in feature of the drone, and through the tracking, the drone can fly autonomously to keep the object visible at a relatively static position in the view. After the tracking, we can crop out the navigator from the frames of the videos and use them as the raw training dataset.

\begin{figure}[t]
    \setlength{\belowcaptionskip}{-0.25cm}   
    \includegraphics[width=0.98\linewidth]{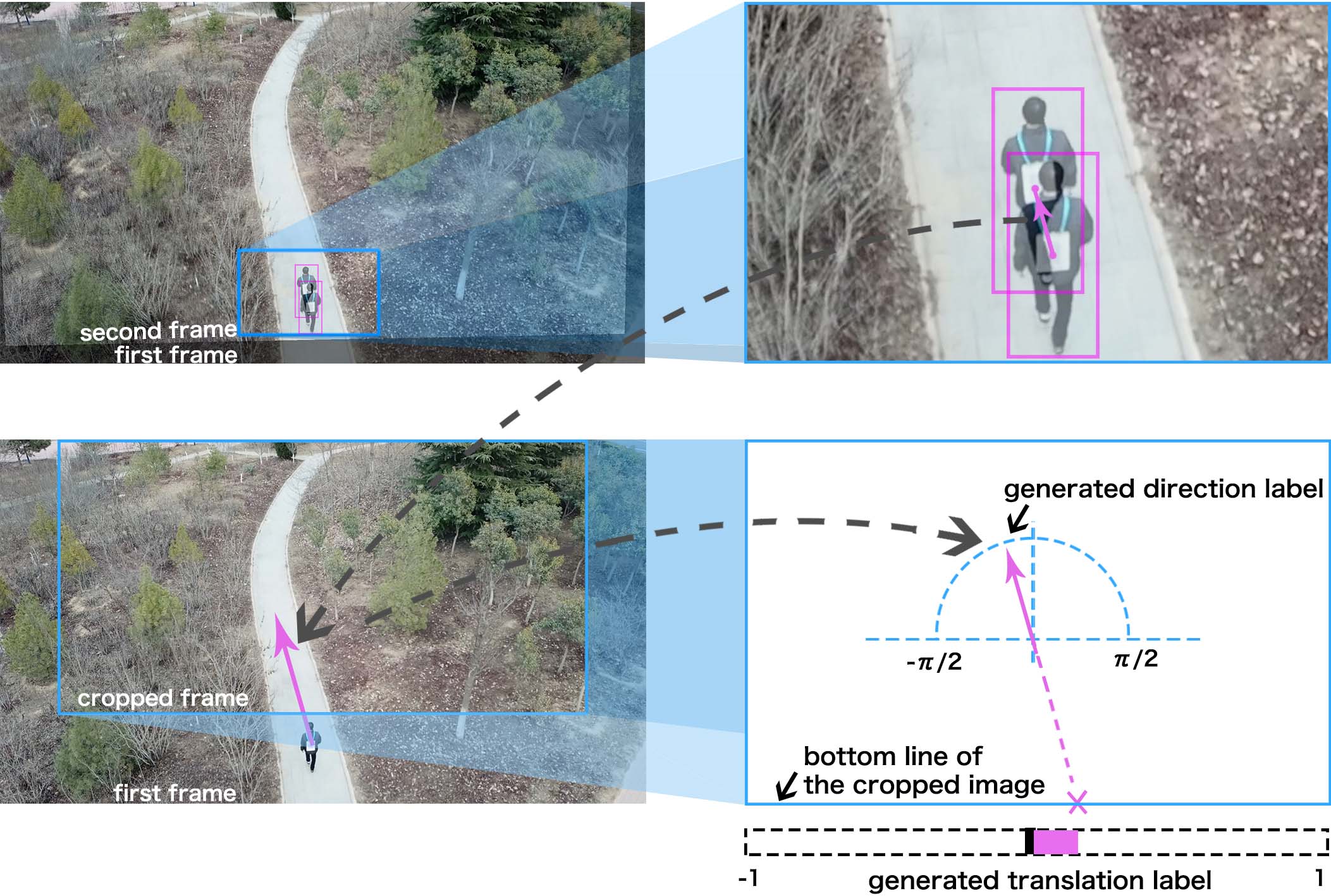}
    \vspace{-4pt}
    \caption{\footnotesize {\bf Video frame auto-labeling.} The top right image is zoomed from the top left image and the navigator's bounding boxes tells the navigator's moving direction and the translation. The two bottom images illustrate the rationale of our auto-labeling algorithm.}
    \label{fig:match}
    \vspace{-6pt}
\end{figure}

{\it Dataset Annotation:} After the observation process, we developed an auto-labeling method based on the inter-frame geometric consistency within the raw videos, which avoids the time-consuming manual annotation and provides a real-valued label for each frame. 
We obtain the labels by computing the translation and direction movements of the navigator using the geometric consistency constraint between successive frames, as the navigator captured in the video is constantly moving along the road. 

The proposed algorithm is given in detail in Algorithm 1 and Fig.~\ref{fig:match}. We first manually select the navigator in the first frame of the video $\textit{B}_0$. Then, we employ a CSRT tracker \cite{lukezic2017discriminative} to track the navigator between the two adjacent frames and get the navigator's position in the next frame $\textit{B}_1$. Since two successive frames within the video are subject to the geometric consistency of a perspective transformation, we then detect and match the feature points between them using SIFT \cite{lowe2004distinctive}. Next, we employ the RANSAC algorithm to estimate a homography matrix $H$ based on the matched feature points to transform the second frame by timing it with $H$ so that it can overlap with the first frame as shown in Fig.~\ref{fig:match}. We compute the center of the bounding box showing the navigator position in the second frame, and also transform it geometrically subject to $H$ to obtain $\tilde{B'}_{i+1}$. We also calculate the center of the bounding box in the first frame $\tilde{B}_{i}$. Finally, by drawing a line connecting the former and latter bounding box centers, we can get the direction label $\textit{D}_{i}$ from the slope of the line, and the point where this line intersects with the bottom line of the cropped image is regarded as the translation label $\textit{T}_{i}$.

 {Our Patrol Dataset includes a variety of observation angles. It refers to the angle between the drone's heading direction and the road direction. Consequently, it helps to overcome the common problem in IL that the distribution of states the expert encounters does not cover all the states the agent encounters \cite{ross2011reduction}. The variety of observation angles are caused by the imperfect drone tracking performance naturally. When the navigator is demonstrating patrolling on a curving road, the drone is likely to a have time-lag for turning, which adds new observation angles. Additionally, the data augmentation we implemented such as flipping, scaling and cropping also help to enrich our dataset.}

\begin{figure*}[t]
    \centering
    \setlength{\belowcaptionskip}{-0.30cm}   
    \includegraphics[width=0.89\textwidth]{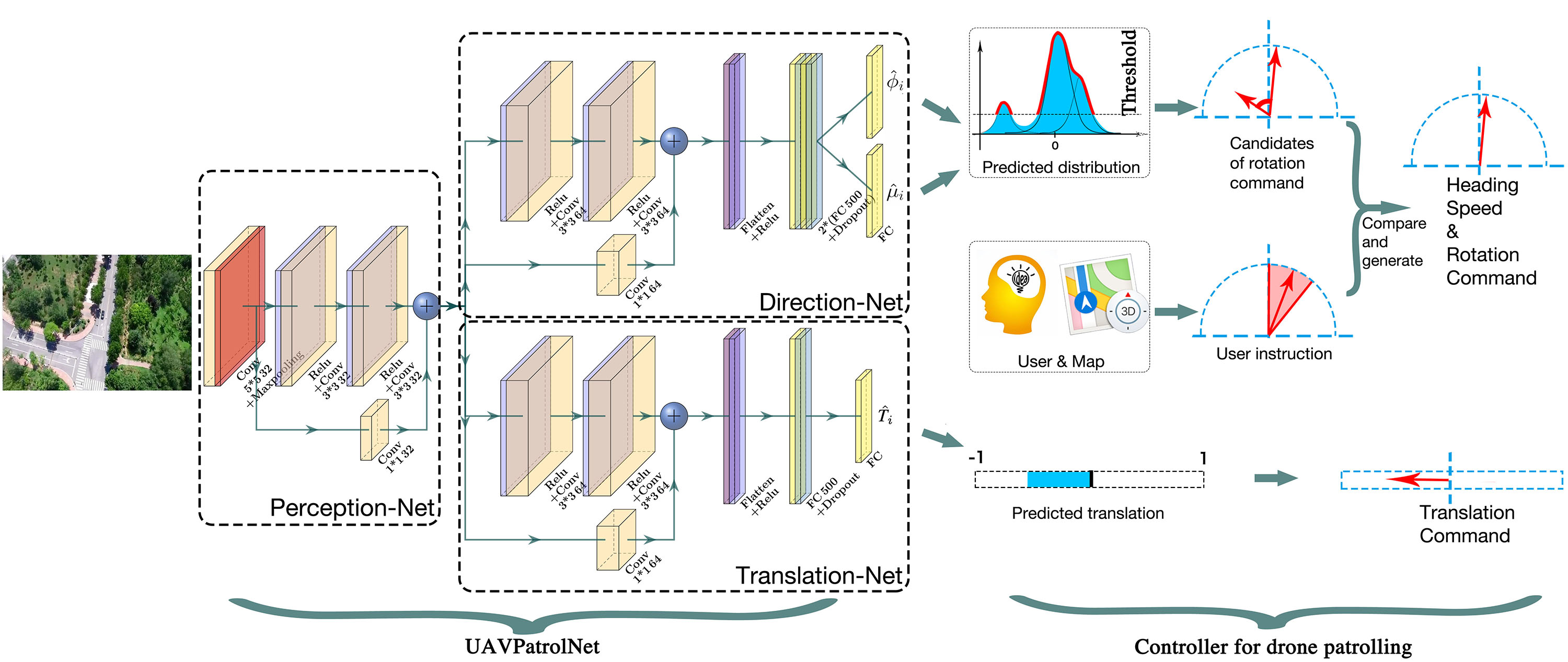} 
    \vspace{-1pt}
    \caption{\footnotesize{{\bf The UAVPatrolNet (left) and the controller (right).
    } The UAVPatrolNet, introduced in Section III.B, has two sets of outputs. The first set of output, $\hat{\boldsymbol{\phi}}, \hat{\boldsymbol{\mu}}$, are the parameters for the predicted probability distribution related to the road direction and the heading speed. The second set of output directly predicts the translation needed for patrolling, which is proportional to the translation command sent to drone. The controller, introduced in Section III.C, uses the outputs of the UAVPatrolNet and optional user instruction to generate the heading speed command $\mathcal{S}$, rotation command $\mathcal{D}$, and translation command $\mathcal{T}$.}} 
    \vspace{-7pt}
    \label{fig:net}
\end{figure*}

\subsection{UAVPatrolNet for Drone Patrolling}

We propose a novel neural network, namely UAVPatrolNet, as shown in Fig.~\ref{fig:net}, to learn human's patrolling behavior such as lane keeping and proper turning at a crossroad. Our UAVPatrolNet is composed of three sub-networks, i.e. Perception-Net, Direction-Net and Translation-Net. 

Perception-Net contains one Resnet V2 block \cite{he2016identity} used to extract features from the input image such as the road position. Direction-Net is a newly designed Mixture Density Network (MDN) aiming to generate the parameters of a Gaussian Mixture Model (GMM) which is defined as the probability distribution over the road direction(s):
\begin{equation}
    P(\boldsymbol{x}|(\boldsymbol{\hat{\phi}}, \boldsymbol{\hat{\mu}}, \boldsymbol{\hat{\sigma}})) = \sum\limits_{i=1}^n \hat{\phi_{i}} \mathcal{N}(\hat{\boldsymbol{\mu_{i}}},\hat{\boldsymbol\sigma_{i}}^2),
    \label{equation_1}
\end{equation}
where $\boldsymbol{x}\in(-1, 1)$ corresponds to the normalized patrolling direction of $(-\pi/2, \pi/2)$. Note that a U-turn with a turning angle larger than $\pi/2$ can be done as a combination of two or three turns with each of the turning angle smaller than $\pi/2$. {\it P(x)} denotes how likely this direction is suitable for the patrol. $\hat{\boldsymbol{\phi}}, \hat{\boldsymbol{\mu}},\hat{\boldsymbol{\sigma}}$ are parameters for the {\it i}-th (${\it i}\leq{\it n}$) components of each GMM, representing the mixing coefficient, mean and variance respectively. We set $n=3$, meaning that we let our model be able to detect up to three different road directions at one time. Also, training the network to output suitable $\hat{\boldsymbol{\sigma}}$ is difficult. To make sure the output $\hat{\boldsymbol{\sigma}}$ is neither too big nor too small, some researchers applied element-wise sigmoid function \cite{choi2018uncertainty}, some use per-component penalty on $\hat{\boldsymbol{\sigma}}$ \cite{amini2019variational}. Since in our case, the variance of each Gaussian function does not matter, we simplify and accelerate the training by fixing the value of $\hat{\boldsymbol{\sigma}}$. Therefore, for each input image {\it I}, our UAVPatrolNet has 2$n$ outputs of GMM parameters. Direction-Net has one Resnet V2 blocks, a ReLU layer and three fully connected layers rather than ResNet-18 architecture with more layers as we find that MDN with fewer layers yields higher accuracy in the drone patrolling task, and using fewer layers also helps to accelerate the training. The major benefit of using MDN is that we are able to train a model that can output the predicted distribution of one or multiple road directions by using a dataset where each training image has only one label of road direction, even if there are multiple road directions within it. Translation-Net includes another residual block, a ReLU layer and two fully connected layers with a dropout of 0.5. It is designed to predict the translation, $\hat{T_i} \in (-1,1)$, needed for keeping the drone on the road in the image taken by its onboard camera. The reason for using less and smaller full connected layers in Translation-Net is that, in practice, the translation output tends to overfit when more fully connected layers are applied.

 %
 
There are two loss functions employed for training. First, the loss function for the Direction-Net's output is the standard negative logarithm of the likelihood:
\begin{equation}
    \begin{split}
      &\mathcal{L}(D=\{\boldsymbol{x}_{1}, \boldsymbol{x}_{2} ... \boldsymbol{x}_{N}\}, ( \boldsymbol{\hat{\phi}}, \boldsymbol{\hat{\mu}},\boldsymbol{\hat{\sigma}}) ) \\
         = &-\sum\limits_{j=1}^N \log \it{P}_{j}(\boldsymbol{x}_{j}| (\boldsymbol{\hat{\phi}}, \boldsymbol{\hat{\mu}}, \boldsymbol{\hat{\sigma}})),
    \end{split}
    \label{equation_2}
\end{equation}
where $D$ is the annotated direction labels of the training dataset, $N$ is the total number of the samples in the dataset, and $( \hat{\boldsymbol{\phi}}, \hat{\boldsymbol{\mu}},\hat{\boldsymbol{\sigma}})  = \textit{Direction-Net}(\textit{Perception-Net}({ \it I})) $ are the outputs of the Direction-Net. Second, the translation output drone should always be a fixed number and thus we simply train it through a mean-squared error (MSE) loss:
\begin{equation}
\mathcal{L}(\hat{T},T) = \frac{1}{N} \sum_{i=0}^{N}(\hat{T_i},-T_i)^2,
\end{equation}
where $T$ denotes the ground truth translation. 

Overall, our UAVPatrolNet is lightweight and meets the requirements for real-time drone patrol. {It is capable of generating unspecified number of outputs as candidate commands especially when the drone encounters intersections where there are multiple possible road directions. The final control command is selected from the candidate commands by the proposed controller described in the next subsection.}

\subsection{Controller and the Fusion of User Instruction} 

\begin{figure}[t]
    \centering
    \setlength{\abovecaptionskip}{-0.1cm}
    \setlength{\belowcaptionskip}{-0.2cm}   
    \subfigure[] 
    {
    \includegraphics[height=3cm]{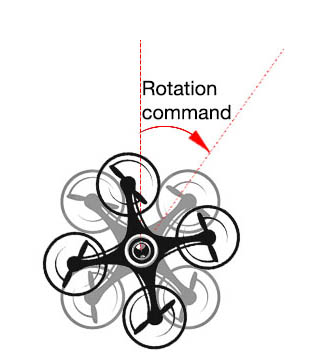}        
    }
    \hspace{0.5cm}
    \subfigure[] 
    {
    \includegraphics[height=3cm]{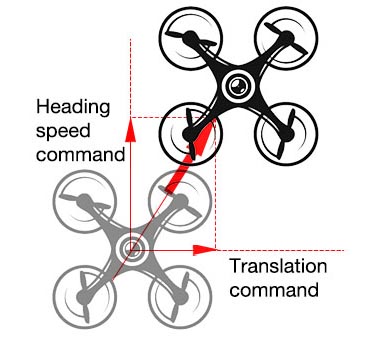}
    }
    
    \caption{\footnotesize {\bf Two ways of control.} (a) Subjected to the rotation command, the drone spins around a point, which is the first mode of the movement. (b)  The translation command lets the drone moving while keeping its head direction unchanged, which is the second mode of the movement.}
    \label{fig:movement}
\end{figure}

While $\hat{\boldsymbol{\phi}}, \hat{\boldsymbol{\mu}},\hat{\boldsymbol{\sigma}}$ output by the proposed UAVPatrolNet are sufficient to generate control commands that enable an automatic lane-keeping patrol, in some cases it also has to handle user-defined tasks. Thus we specifically design a controller that provides the extra functionality that optional user instructions can be fused with the candidate commands derived from the outputs of the UAVPatrolNet.  

The proposed controller needs to generate two outputs in order to control the drone. They are the translation command and rotation command corresponding to the drone's two modes of movement as shown in Fig.~\ref{fig:movement}. The inputs are the predicted parameters of probability distribution over the road direction, the predicted road center position and the user instruction if available, where first two inputs come from the UAVPatrolNet and are used to generate multiple candidate commands of rotation and one translation commands. The user instruction is involved in the process of computing rotation command because when flying above crossroads, there could be multiple road directions observed. When the user instruction is given, the controller will choose the final translation command that is most similar to the user instruction from the candidate translation commands. {When there is no input from the user, one candidate command closest to zero will be chosen by default.} 

The process of our controller generating control commands is as follows: At each control period, the controller first takes in the outputs of the UAVPatrolNet's MDN part, i.e. the parameters of the GMM and generates the mixture model. The mixture model represents the probability distribution as demonstrated in Fig.~\ref{fig:net}. The mid-point along the $x$-axis of each probability distribution section larger than a predefined threshold will be selected as the candidates of patrolling rotation ${D_{\textit{CAND}}(j), \in(-\pi/2, \pi/2)}, j=1,2,...,C$, where $C$ is the total number of candidates. Then, the controller uses the user instruction $D_u$ (processed to the same form as the candidate rotation) to choose the final rotation command:
\begin{equation}
\begin{aligned}
  &\mathcal{D} = \alpha D_{\textit{CAND}}(j), \\
  &\textrm{where} \quad j = \argmin_{j} ||D_{\textit{CAND}}(j)-D_u||.
	\end{aligned}
\end{equation}
$\mathcal{D} \in(-\pi/2, \pi/2) $ is the final rotation command with a unit of $rad/s$. $\alpha$ is a constant that converts the unit of $rad$ to the unit of $rad/s$. If the user instruction is not provided, the candidate command closest to zero is output.

At the same time, we also set the drone's heading speed $\mathcal{S} \in (-10, 10)$ with a unit of $m/s$ according to the probability of the candidate command. As a result, if the rotation command corresponds to a high probability of accessible road direction, the drone will move fast toward that direction. Otherwise, the drone will slow down its speed. Finally, the controller generate the translation command $\mathcal{T} \in(-3, 3)$ with a unit of $m/s$ by scaling the UAVPatrolNet's translation output $\hat{T}$. The two control commands are implemented by drone simultaneously. Due to translation command, the drone will keep the road in the center of the onboard camera's view and the rotation command will make the drone turn to the direction where the road leads. In this way, drone's flexibility can be fully explored and a better control effect is achieved.

\begin{figure}[t]
    \centering
    \setlength{\abovecaptionskip}{-0.1cm}
    \setlength{\belowcaptionskip}{-0.3cm}   
    \subfigure[Hardware] 
    {
        \includegraphics[width=0.8\linewidth, height = 4.3cm]{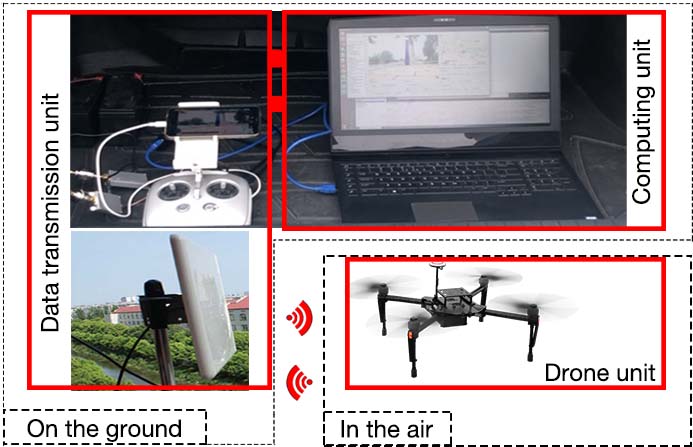}
        \label{fig:system}
    }
    
    \subfigure[Interface] 
    {
    
        \includegraphics[width=0.8\linewidth, height = 4.2cm]{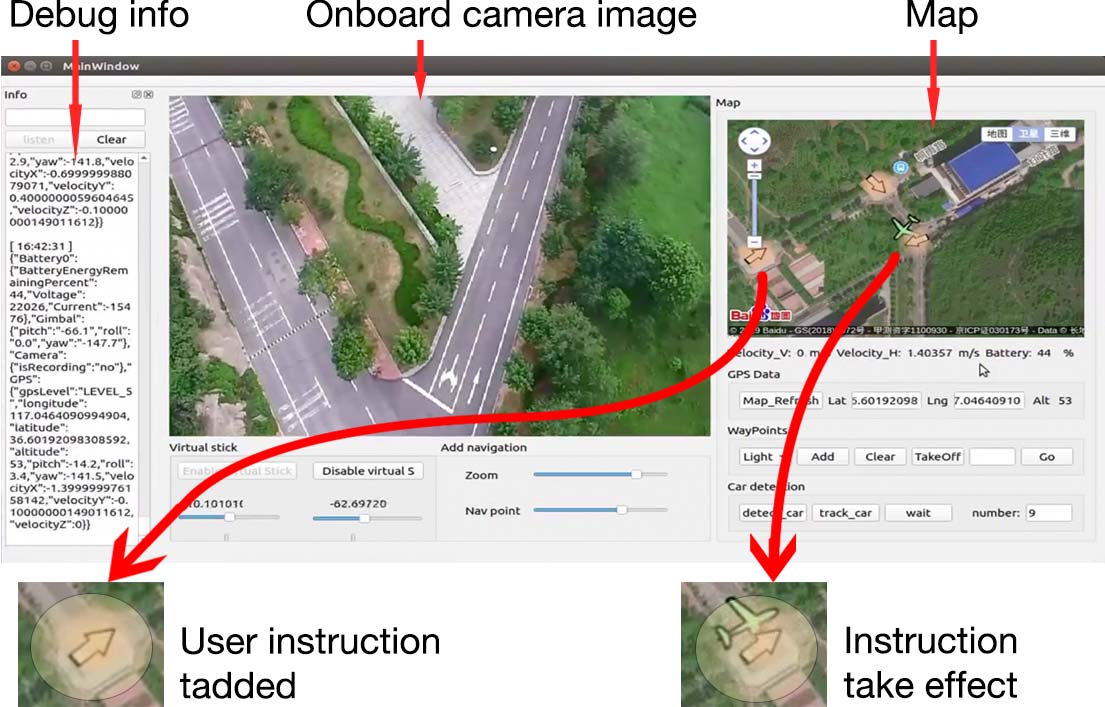}
        \label{fig:interface}
        
    }
    
    \caption{\footnotesize {\bf Overview of the holistic system.} (a) shows the hardware system with three parts. (b) shows the user interface. When the user clicks the map and the button to add instruction on the interface, an arrow icon will be added to the map. The user instruction will take effect when the drone's approximate position gets into the effect zone set with the user interface. }

\end{figure}

\section{System Setup}
We aim to develop a drone system that can navigate autonomously and also perform specific tasks simultaneously if necessary. A high-performance computing unit is thus required. Since the drone's carrying capacity is limited, we choose to use off-board control strategy, which means that the unit computing and generating control commands is not carried by the drone. As a result, the real system we designed is composed of three separated parts as shown in Fig.~\ref{fig:system}: drone unit, computing unit, and data transmission unit. 

First, we use DJI Matrice 100 as the drone unit. It mainly consists of an onboard camera with gimbal control, a DJI N1 flight controller and the motor module. Second, we choose a laptop with an NVIDIA GTX970 graphic card as the computing unit. In the experiments, images taken by the Matrice 100's onboard camera will be sent to the laptop and computed by the proposed neural network. {The input of our UAVPatrolNet are images taken by the drone's onboard camera resized to $400\times100$, which benefits the efficiency while preserving the geometry content and features of the input images. As a result, the frame rate is 24.1 fps on the laptop 
. The output final control commands are used for controlling the drone unit, and in order to save the transmission bandwidth, we set the control rate at 10hz.} For the data transmission unit, the major part is the Matrice 100's remote controller. We customized an active antenna so that the controller can connect with the drone at a much longer distance. Also, we developed an Android app using DJI Mobile APK. The app is linked to the remote controller by wire and at the same time, can exchange data with the computing unit through Internet TCP socket. By this way, computing unit is able to communicate with the drone unit. 

In order to enable the user to give instructions and make our drone patrolling system easy to use in real world, we developed a user interface as shown in Fig.~\ref{fig:interface}. The interface can visualize the system status and enable the user to give sparse instructions to the drone system. The map showed in the interface can only roughly represent the drone position, which means the actual drone position falls within a certain range of the presented drone position. In the same way, the instructions from the user also have a certain effective range. The user instructions can be given either before the patrol task or during the patrol task. When the distance between the approximate drone position and the instructed point is smaller than 1 km, i.e. within the effect zone, the instruction will take effect and help the controller select the control command from the candidate commands. 

\section{Experimental Results}
In this section, we first conduct an experiment in a simulated environment to test the accuracy of the auto-labeling algorithm that we used for creating the annotations of the observed video dataset. Then, we compare our UAVPatrolNet with several baseline models in terms of prediction accuracy as well as flexibility. Finally, we let our drone patrol in the real world autonomously and evaluate its performance.

\begin{figure}[t]
    \centering
    \setlength{\abovecaptionskip}{-0.1cm}
    \setlength{\belowcaptionskip}{-0.2cm}   

    \subfigure[Simulation environment setup]
    {
        \includegraphics[width=0.465\linewidth,height=2.7cm]{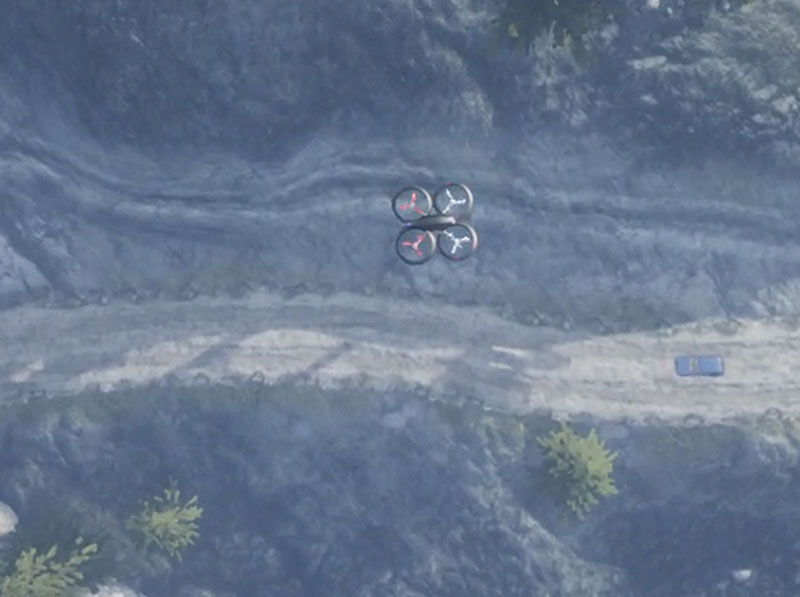}
        \label{fig:airsim}
    }
    \subfigure[Numerical evaluation result]
    {
        \includegraphics[width=0.465\linewidth,height=2.7cm]{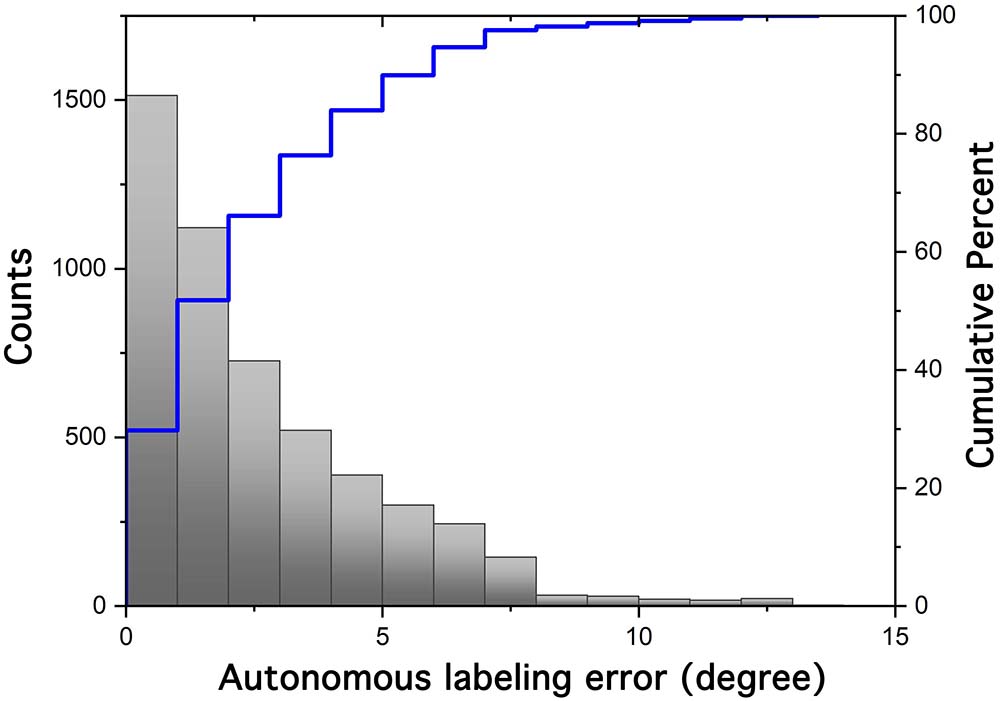}
        \label{fig:accuracy_in_airsim}
    }
    \caption{\footnotesize {\bf Auto-labeling results in Airsim simulator.} (a) shows the scene in our simulator. In (b), the rectangular bars correspond to the numbers of the testing samples and the blue line is their cumulative percentage.}
    
\end{figure}
\subsection{Data Auto-labeling Accuracy}

In order to quantitatively evaluate our autonomous data collection and auto-labeling method, we test it in Airsim \cite{shah2018airsim} which is a drone simulator built on the Unreal Engine. The simulated drone in the simulator uses a PX4 flight controller. Through the Software-in-Loop (SITL) mode, we can control the simulated PX4 drone as in the real world. We employ a PX4 drone tracking algorithm and enable the simulated drone to track the virtual navigator just as the DJI Mavic drone in the real world. Since we can extract the accurate position of the navigator in the simulator, we can quantitatively compare the difference between the generated labels and the ground truth of the navigator's motion. 

We first set up the simulated environment shown in Fig.~\ref{fig:airsim}. The navigator in the simulated environment is a car moving along the road with a fixed route. The drone in the air will track the car based only on visual data, which is exactly the same as the setup in the real world. We save the images taken by the onboard camera and use our auto-labeling method to generate labels for them. After that, we compute the absolute difference between the ground truth direction labels and the generated direction labels as autonomous labeling error, because the error of translation label is determined by direction label error. As shown in the Fig.~\ref{fig:accuracy_in_airsim}, the labeling error of translation is less than 5 degrees for more than 70\% of the data, indicating high accuracy. It is noteworthy that our codes for auto-labelling as well as our Patrol Dataset have been released and made publicly available at the website mentioned in the Abstract.

\subsection{UAVPatrolNet Evaluation}

We train our UAVPatrolNet with our Patrol Dataset containing around $30,000$ images. {Each image is collected and labeled autonomously at the resolution of 1080p, which will be cropped and resized before being fed into the network.} After a 36-hour training on an NVIDIA 1080Ti GPU, we evaluate our model by comparing its prediction accuracy on the testing set of our Patrol Dataset, including 643 images, with a random output and two highly cited models designed for drone patrolling: Dronet \cite{loquercio2018dronet} and TrailNet \cite{smolyanskiy2017toward}.

{In the evaluation, we first evaluate the root mean square error of both the patrolling direction output and the translation output (RMSE of $\hat{D}$ and RMSE of $\hat{T}$). Then,} prediction accuracy of the patrolling direction output ($\hat{D}$ Acc.) is defined as the percentage of the sample images in the testing set assigned with a correct label of predicted direction. A predicted label of a testing image is correct as long as the difference between predicted patrolling direction and the ground truth label is less than a threshold which is set to $\pi/12$. The accuracy of the translation output ($\hat{T}$ Acc.) is defined as the percentage of the sample images in the testing set that are assigned with a correct label of predicted translation. Similarly, we allow a tolerance of 0.2 with regard to the ground truth. 
Also, we measure the difference between the predicted label and the ground truth label via the L2 norm and calculate the standard deviation of the differences over the direction and translation prediction (SD of $\hat{D}$ Err. and SD of $\hat{T}$ Err.) for the entire testing set. TABLE~\ref{table_1} shows the comparative results where our UAVPatrolNet outperforms the competitors in terms of both prediction accuracy and standard deviation, indicating its accurate and stable navigation capability. {Furthermore, we split our test dataset into two subsets based on the widths of the roads and evaluate the RMSEs of $\hat{D}$ and $\hat{T}$. The subset containing the images of roads with widths larger than 15\% of the image width has 0.11 RMSE of $\hat{D}$ and 0.08 RMSE of $\hat{T}$. The other subset, with road widths smaller than 15\% image width has 0.17 RMSE of $\hat{D}$ and 0.12 RMSE of $\hat{T}$. Therefore, our network performs well on scenes with different road widths, and especially better when the road can be captured at a width larger than 15\% of the image width.}

\begin{figure}
    \centering
    \setlength{\abovecaptionskip}{-0.1cm}
    \setlength{\belowcaptionskip}{-0.25cm}   
    \subfigure[] 
    {
        \includegraphics[height=2.84cm]{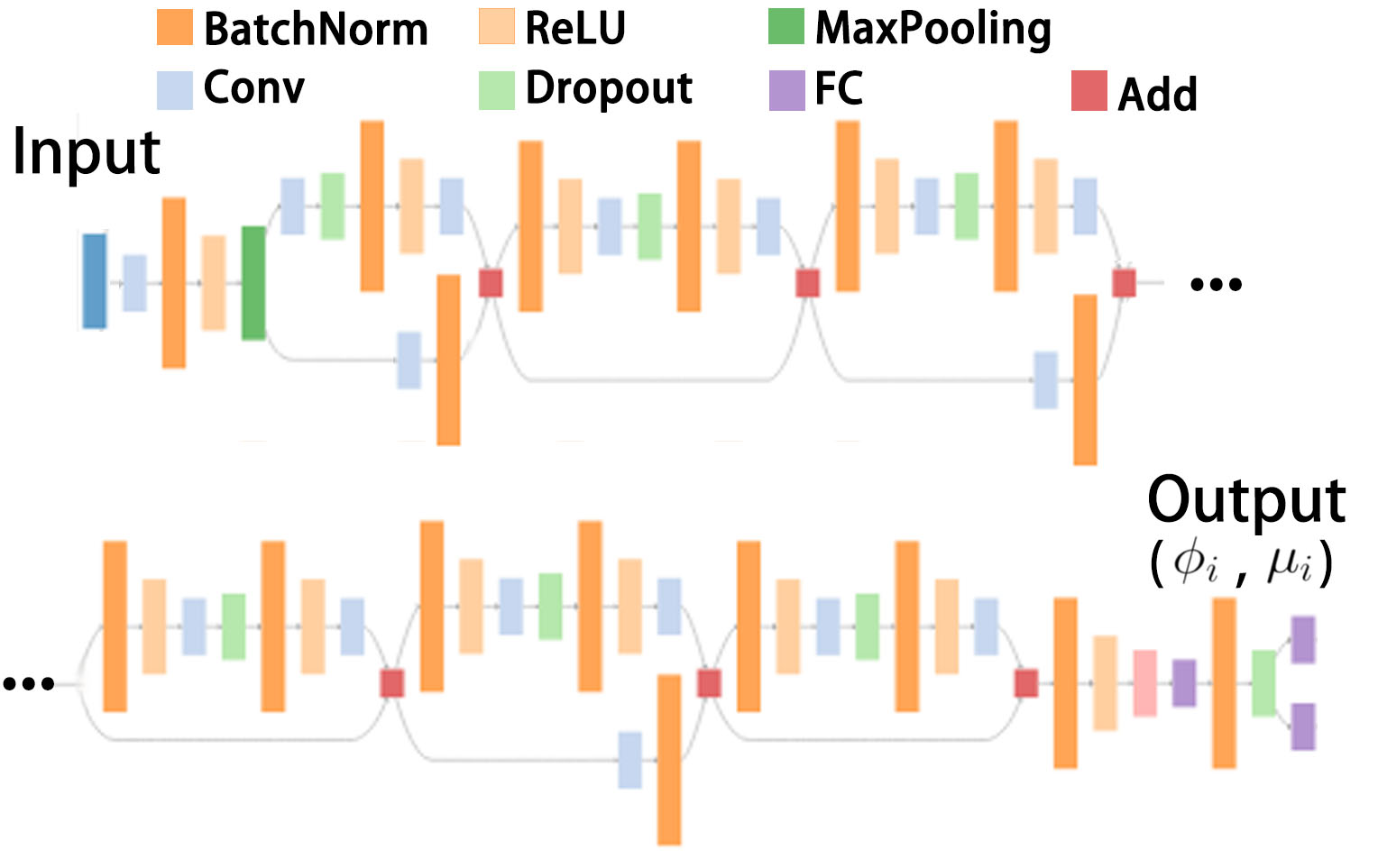}
        \label{fig:compare_net}
    }
    \subfigure[] 
    {
        \includegraphics[height=3.0cm]{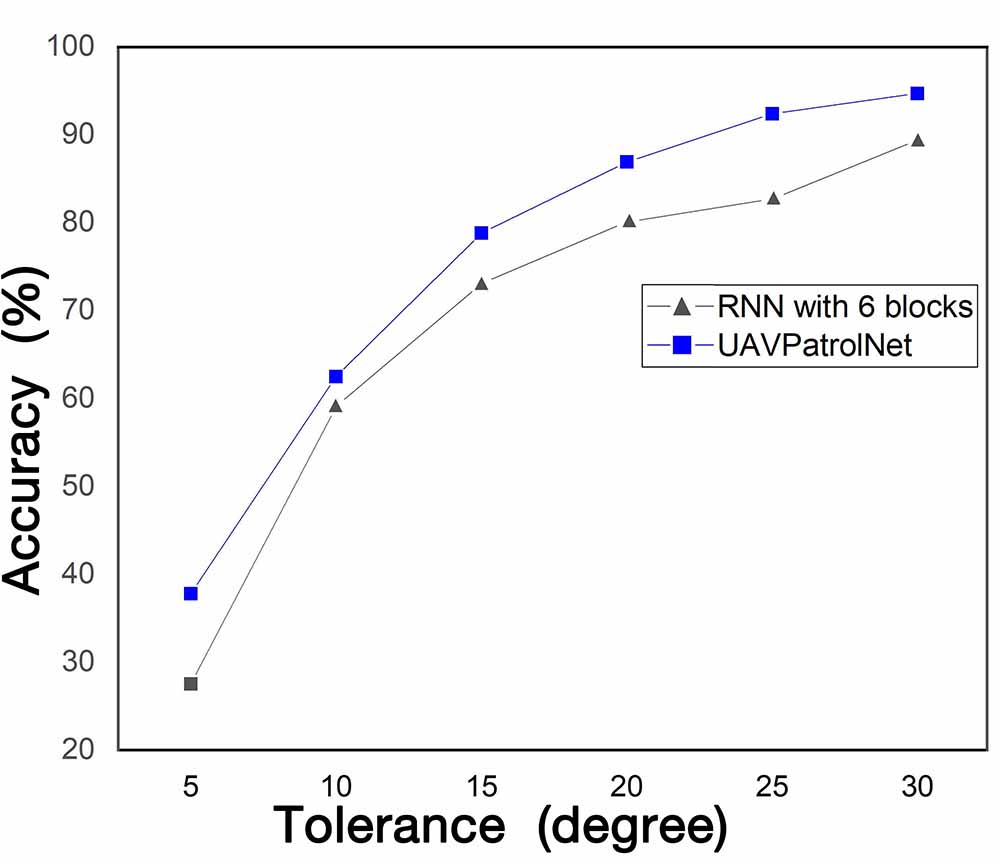}
        \label{fig:compare_net_results}
    }
    
    \caption{\footnotesize{(a) Accuracy comparison between the proposed UAVPatrolNet and the neural network with 6 residual blocks. (b) Structure of the neural network with 6 residual blocks} }
    \label{fig:compare}

\end{figure}

In our UAVPatrolNet, the Direction-Net and the Translation-Net shares the output of Perception-Net. Theoretically, the training of the Perception-Net is more difficult as the weights of it are influenced by two losses that come from the two sets of the output. Compared to our UAVPatrolNet, the normal residual neural networks use deeper network structures. To comparatively demonstrate that the design of the UAVPatrolNet is reasonable, we design another neural network with 6 residual blocks as illustrated in Fig.~\ref{fig:compare_net}. It is trained using the same data with the same number of epochs and only outputs the predicted parameters of the GMM. By comparing its accuracy with our UAVPatrolNet in Fig.~\ref{fig:compare_net_results}, we prove that the shared layers in the UAVPatrolNet do not decrease the accuracy of the prediction.

{}

\begin{table}[h]
\scriptsize{
    \setlength{\abovecaptionskip}{-0.1cm}
    \setlength{\belowcaptionskip}{-0.1cm}
    
    \caption{\small{{\bf Evaluation on Patrol Dataset (testing set).}}}
    \label{table_1}
    \begin{center}
    \begin{tabular}{|c||c|c|c|c|}
    \hline
    &Random &  TrailNet & DroNet & UAVPatrolNet  \\
    
    \hline
    
    RMSE of $\hat{D}$ & $ 0.41 \pm 0.03$ & $ 0.18 $ & $ 0.19 $ & $ \textbf{0.15} $\\
    \hline
    
    RMSE of $\hat{T}$ & $ 0.63 \pm 0.04$ & $ 0.13 $ & $ - $ & $ \textbf{0.11} $\\
    \hline
    
    $\hat{D}$ Acc. & $29.5\pm2.0\%$ & $ 76.7 \%$ & $ 69.4 \%$ & $\textbf{78.8}\%$ \\
    \hline
    $\hat{T}$ Acc. & $21.4\pm1\%$  & $ 89.0 \%$ & $ - $ & $\textbf{93.9}\%$\\
    \hline
    SD of $\hat{D}$ Err. & $ 0.46 \pm 0.01 $ & $ 0.17 $ & $0.26 $ & $ \textbf{0.15} $\\
    \hline
    SD of $\hat{T}$ Err. & $ 0.62 \pm 0.01$ & $ 0.12 $ & $ - $ & $ \textbf{0.12} $\\
    \hline
    
    \end{tabular}
    \vspace{-0.5cm}
    \end{center}
}
\end{table}

\begin{figure}[t]
    \centering
    \setlength{\abovecaptionskip}{-0.1cm}
    \setlength{\belowcaptionskip}{-0.25cm}   
    \subfigure[Flying route (marked in red)] 
    {
        \includegraphics[height=2.4cm]{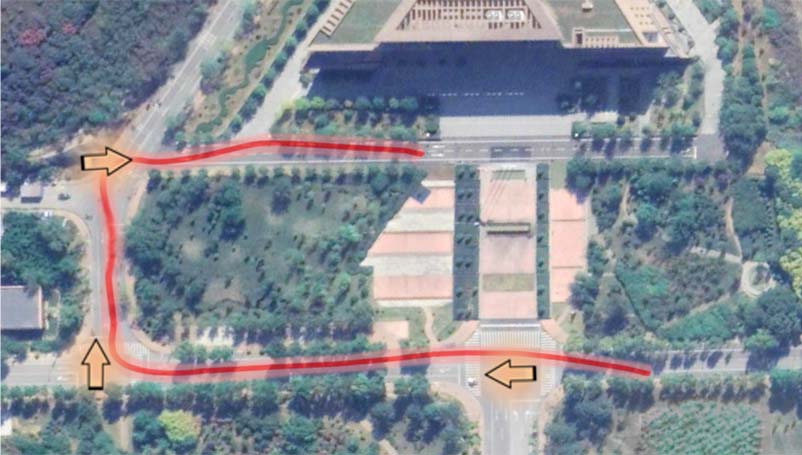}
    }
    \subfigure[Serialized camera views] 
    {
        \includegraphics[height=2.45cm]{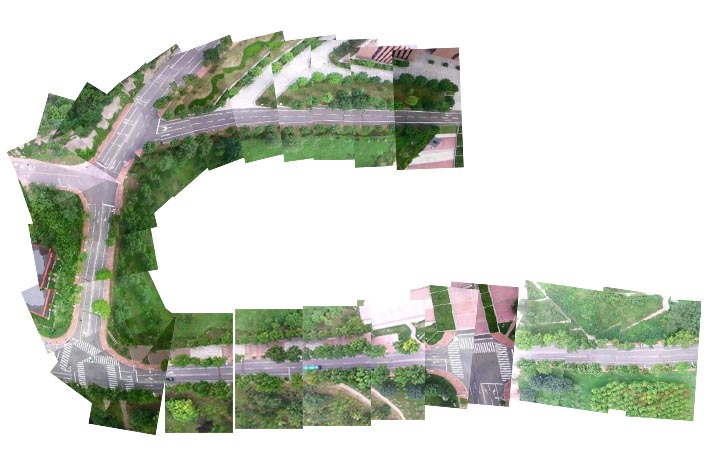}
    }
    \caption{\footnotesize{\bf Flying route and serialized images from the onboard camera.} We manually correct the GPS data in order to generate the flying route in (a) and stack the frames from the onboard camera during the patrol in (b). The average flying altitude is 60 meters.}
    \label{fig:testfly_in_real}
\end{figure}

\begin{figure*}[th]
    \centering
    \setlength{\abovecaptionskip}{-0.1cm}
    \setlength{\belowcaptionskip}{-0.2cm}   
    \subfigure[Successful results] 
    {
        \centering
        \begin{minipage}[b]{0.18\linewidth}
            \centering
            \includegraphics[width=1\linewidth,height=2cm]{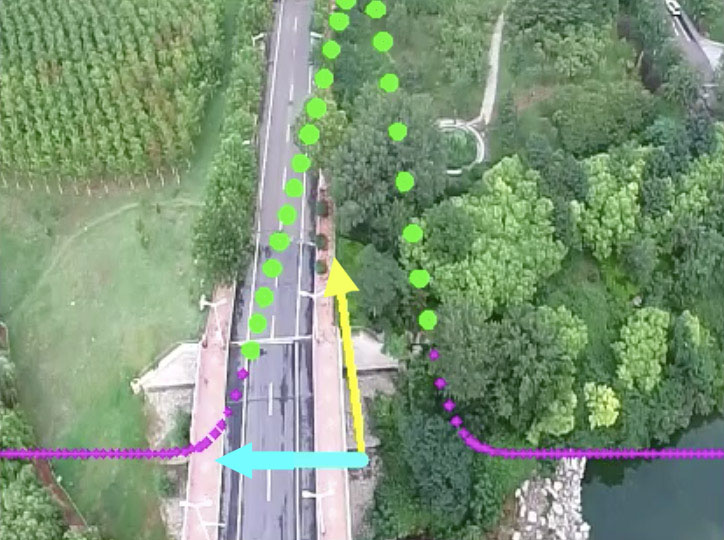}\vspace{0.1cm}
            \includegraphics[width=1\linewidth,height=2cm]{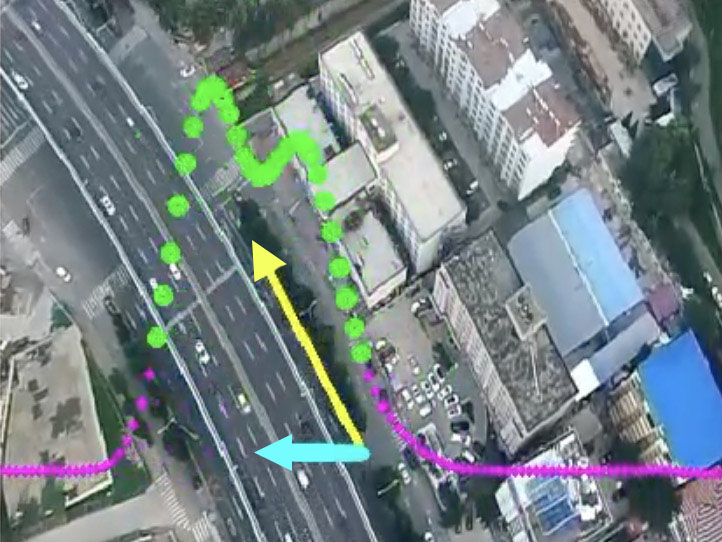}
        \end{minipage}
        
        \begin{minipage}[b]{0.18\linewidth}
            \centering
            \includegraphics[width=1\linewidth,height=2cm]{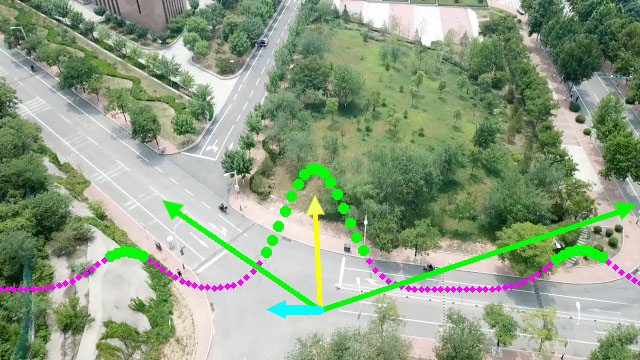}\vspace{0.1cm}
            \includegraphics[width=1\linewidth,height=2cm]{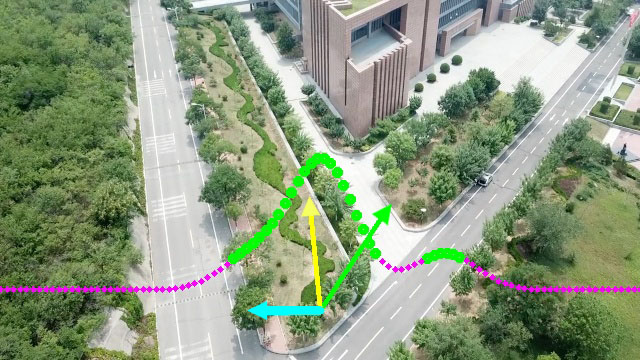}
        \end{minipage}
    \label{fig:successful_and_in}
    }\hspace{0.02cm}
    \subfigure[Less successful results] 
    {
        \begin{minipage}[b]{0.18\linewidth}
            \centering
            \includegraphics[width=1\linewidth,height=2cm]{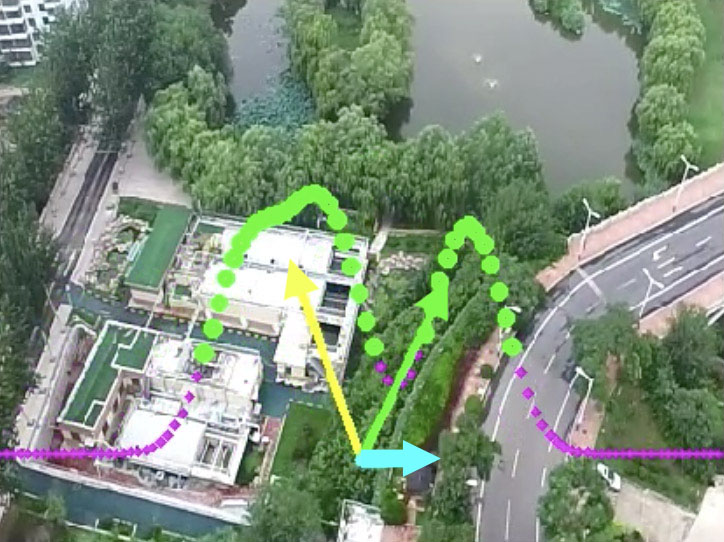}\vspace{0.1cm}
            \includegraphics[width=1\linewidth,height=2cm]{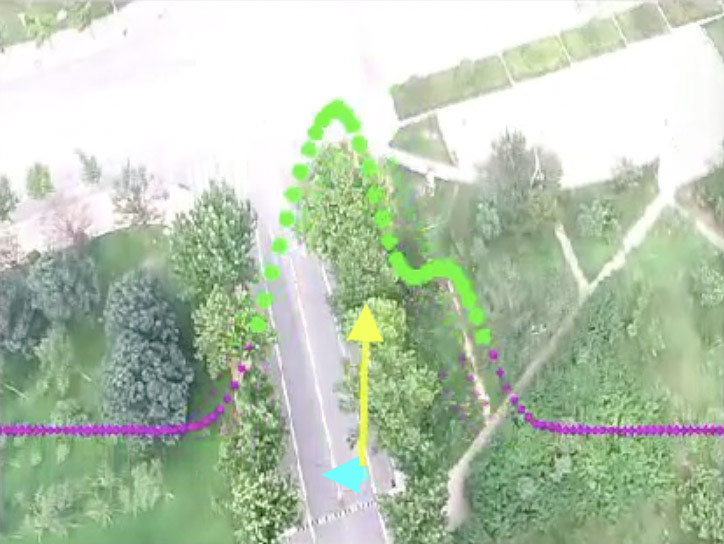}
        \end{minipage}
    \label{fig:insuccessful}
    }\hspace{0.02cm}
    \subfigure[TrailNet results] 
    {
        \begin{minipage}[b]{0.18\linewidth}
            \centering
            \includegraphics[width=1\linewidth,height=2cm]{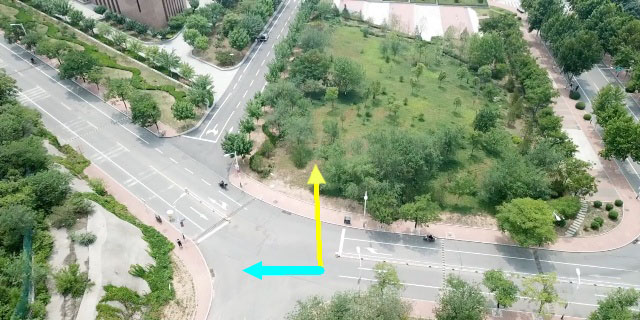}\vspace{0.1cm}
            \includegraphics[width=1\linewidth,height=2cm]{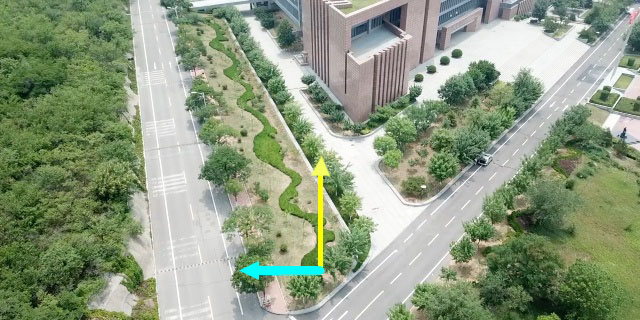}
        \end{minipage}
    \label{fig:TrailN}
    }\hspace{0.02cm}
    \subfigure[DroNet results] 
    {
        \begin{minipage}[b]{0.18\linewidth}
            \centering
            \includegraphics[width=1\linewidth,height=2cm]{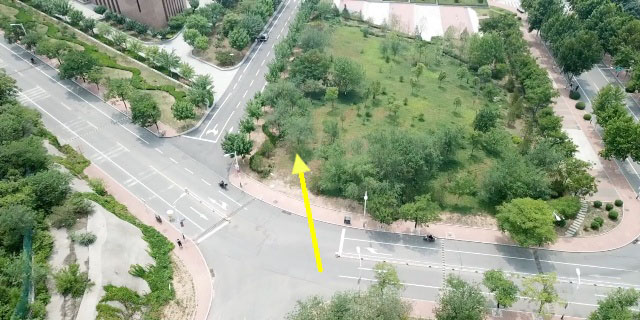}\vspace{0.1cm}
            \includegraphics[width=1\linewidth,height=2cm]{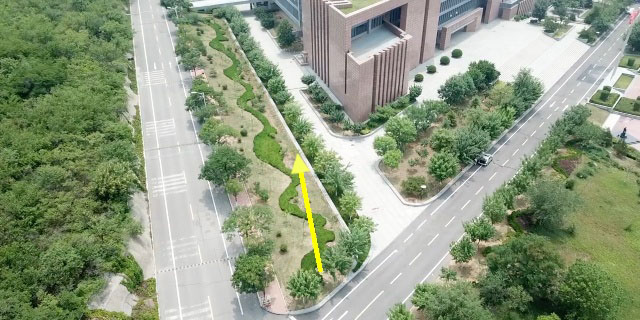}
        \end{minipage}
    \label{fig:droN}
    }
    \caption{\footnotesize {\bf Images taken by the drone's onboard camera with testing results.} The dotted lines denote the predicted probability distribution where sections higher than a threshold  are colored in green and and others are in purple. Each yellow or green line corresponds to a group of contiguous green dots representing a possible road direction. The yellow line denotes the direction finally selected by the drone. The horizontal cyan line is the translation output.}
    
\end{figure*}

\subsection{Patrolling with Autonomous Navigation System}

We test our drone autonomous navigation system for patrolling in various real-world environments, including highway, campus driveway and walking path between buildings.

One video of the patrolling results is sampled and shown in serialized frames (see Fig.~\ref{fig:testfly_in_real}). {The flying route length is about 1km, and there are three user instructions provided during the flight. }Although some frames are overexposed, it does not make the drone drift off its patrol route, as the heading speed of the drone is reduced by the relatively flatten distribution of predicted probabilities. We evaluate the accuracy of the patrol route by scoring the patrol accuracy at each frame. Specifically, we manually create a ground truth of road direction and road center position. The direction scores of each frame are inversely proportional to the difference between the road direction and the forward direction. The translation score is inversely proportional to the distance between the road center and the image center. The direction and translation scores of the automatic patrol route are 86.6\% and 80.7\% respectively, which are comparable with the scores of 88.1\% and 83.3\%, for the pilot with one-month flight experience controlling the drone to patrol in the same route.

Additionally, some successful and less successful test results are shown in Fig.~\ref{fig:successful_and_in} and Fig.~\ref{fig:insuccessful}. The less successful examples show that the UAVPatrolNet could be misled by some patterns or overexposure. However, since the heading speed of the drone is determined by the predicted probability distribution, the drones flies slowly when the UAVPatrolNet is misled. Therefore, those less successful cases have limited impact on the patrol. Last but not least, we compared our UAVPatrolNet with DroNet \cite{loquercio2018dronet} and TrailNet \cite{smolyanskiy2017toward} at scenarios of crossroads. As shown in Fig.~\ref{fig:TrailN} and Fig.~\ref{fig:droN}, both competing methods can only generate one patrolling directions which makes the patrol along a desired route impossible. Please refer to the website mentioned in the Abstract for the video demonstrations and source codes of our method.

\section{CONCLUSIONS}

We have designed an IL-based system for automatic drone patrol. It uses UAVPatrolNet to learn human patrolling behaviors recorded in raw videos which are collected and annotated by the system automatically, and then generates control commands based on the outputs of the UAVPatrolNet and optional user instructions. Such a design strategy makes the patrolling controllable and flexible. We carried out extensive experiments to demonstrate the effectiveness of our system and the superiority over two state-of-the-art competitors. The proposed UAVPatrolNet is lightweight and runs in real-time rate. We hope in the future it can be applied to many professional applications like target searching and traffic inspection. 











\bibliographystyle{IEEEtran}
\bibliography{bib}

\end{document}